\documentclass{article}
\usepackage[final, nonatbib]{neurips_like}

\newcommand{\usefinal}{}
\usefinal

\usepackage[utf8]{inputenc} 
\usepackage[T1]{fontenc}    
\usepackage{url}            
\usepackage{booktabs}       
\usepackage{amsfonts}       
\usepackage{nicefrac}       
\usepackage{microtype}      
\usepackage{amsmath,amssymb}

\usepackage{theorem}
\usepackage{algorithm}
\usepackage{algorithmic}
\usepackage[all]{xy}

\newtheorem{theorem}{Theorem}
\newtheorem{proposition}{Proposition}

\RequirePackage{ifpdf}
\ifpdf
  \usepackage[pdftex]{graphicx}
  \usepackage[pdftex]{hyperref}       
\else
  \usepackage[dvipdfmx]{graphicx}
  \usepackage[dvipdfmx]{hyperref}       
\fi
\usepackage{color}

\newcommand{\E}[2]{\mathbb{E}_{#1} \left[
#2
\right]
}
\def\bm#1{\mbox{\boldmath $#1$}}
\def\sbm#1{\mbox{\scriptsize \boldmath $#1$}}
\def\ssbm#1{\mbox{\tiny \boldmath $#1$}}
\DeclareMathOperator*{\argmin}{arg\,min}

\title{
Discriminator optimal transport
}

\author{
  Akinori Tanaka\\
  Mathematical Science Team, 
  RIKEN Center for Advanced Intelligence Project (AIP) \\
  1-4-1 Nihonbashi, Chuo-ku, Tokyo 103-0027, Japan \\
  Interdisciplinary Theoretical and Mathematical Sciences Program (iTHEMS), RIKEN \\
2-1 Hirosawa, Wako, Saitama 351-0198, Japan
\\
  Department of Mathematics, Faculty of Science and Technology,
Keio University\\
3-14-1 Hiyoshi, Kouhoku-ku, Yokohama 223-8522, Japan
\\
 \texttt{akinori.tanaka@riken.jp} \\
}

\begin{document}

\maketitle

\begin{abstract}
Within a broad class of generative adversarial networks, we show that discriminator optimization process increases a lower bound of the dual cost function for the Wasserstein distance between the target distribution $p$ and the generator distribution $p_G$.
It implies that the trained discriminator can approximate optimal transport (OT) from $p_G$ to $p$.
Based on some experiments and a bit of OT theory, we propose discriminator optimal transport (DOT) scheme to improve generated images.
We show that it improves inception score and FID calculated by un-conditional GAN trained by CIFAR-10, STL-10 and a public pre-trained model of conditional GAN trained by ImageNet.
\end{abstract}

\renewcommand{\include}[1]{}
\renewcommand\documentclass[2][]{}
\newcommand{\inputt}[1]{}


%

\documentclass{article}

\newcommand{\inputt}[1]{\input{#1}}
\newcommand{\usefinal}{\usepackage[final]{neurips_2019}}


\maketitle

\begin{abstract}
Within a broad class of generative adversarial networks, we show that discriminator optimization process increases a lower bound of the dual cost function for the Wasserstein distance between the target distribution $p$ and the generator distribution $p_G$.
It implies that the trained discriminator can approximate optimal transport (OT) from $p_G$ to $p$.
Based on some experiments and a bit of OT theory, we propose discriminator optimal transport (DOT) scheme to improve generated images.
We show that it improves inception score and FID calculated by un-conditional GAN trained by CIFAR-10, STL-10 and a public pre-trained model of conditional GAN trained by ImageNet.
\end{abstract}

\section{Introduction}
Generative Adversarial Network (GAN) \cite{goodfellow2014generative} is one of recent promising generative models.
In this context, we prepare two networks, a generator $G$ and a discriminator $D$.
$G$ generates fake samples $G(\bm z)$ from noise $\bm z$ and tries to fool $D$.
$D$ classifies real sample $\bm x$ and fake samples $\bm y = G(\bm z)$.
In the training phase, we update them alternatingly until it reaches to an equilibrium state.
%
In general, however, the training process is unstable and requires tuning of hyperparameters. 
Since from the first successful implementation by convolutional neural nets \cite{radford2015unsupervised},
most literatures concentrate on \textit{how to improve the unstable optimization procedures} including changing objective functions \cite{nowozin2016f, zhao2016energy, arjovsky2017wasserstein, lim2017geometric, unterthiner2017coulomb, bellemare2017cramer}, adding penalty terms \cite{gulrajani2017improved, petzka2017regularization, wei2018improving}, techniques on optimization precesses themselves \cite{metz2016unrolled, salimans2016improved, karras2017progressive, heusel2017gans}, inserting new layers to the network \cite{miyato2018spectral, zhang2018self}, and others we cannot list here completely.

Even if one can make the optimization relatively stable and succeed in getting $G$ around an equilibrium, it sometimes fails to generate meaningful images.
Bad images may include some unwanted structures like unnecessary shadows, strange symbols, and blurred edges of objects.
For example, see generated images surrounded by blue lines in Figure \ref{fig:ot_all_imagenet}.
These problems may be fixed by scaling up the network structure and the optimization process. Generically speaking, however, it needs large scale computational resources, and if one wants to apply GAN to individual tasks by making use of more compact devices, the above problem looks inevitable and crucial.

There is another problem.
In many cases, we discard the trained discriminator $D$ after the training.
This situation is in contrast to other latent space generative models.
For example, variational auto-encoder (VAE) \cite{kingma2013auto} is also composed of two distinct networks, an encoder network and a decoder network.
We can utilize both of them after the training:
the encoder can be used as a data compressor, and the decoder can be regarded as a generator.
Compared to this situation, it sounds wasteful to use only $G$ after the GAN training.

From this viewpoint, it would be natural to ask \textit{how to use trained models $G$ and $D$ efficiently}.
Recent related works in the same spirit are discriminator rejection sampling (DRS) \cite{azadi2018discriminator} and Metropolis-Hastings GAN (MH-GAN) \cite{turner2019metropolis}.
In each case, they use the generator-induced distribution $p_G$ as a proposal distribution, and approximate acceptance ratio of the proposed sample based on the trained $D$.
Intuitively, generated image $\bm y = G(\bm z)$ is accepted if the value $D(\bm y)$ is relatively large, otherwise it is rejected.
They show its theoretical backgrounds, and it actually improve scores on generated images in practice.

In this paper, we try similar but more active approachs, i.e. improving generated image $\bm y = G(\bm z)$ directly by adding $\delta \bm y$ to $\bm y$ such that $D(\bm y + \delta \bm y) > D(\bm y)$ by borrowing idea from the optimal transport (OT) theory.
More concretely, our contributions are
\begin{itemize}
\item Proposal of the discriminator optimal transport (DOT) based on the fact that the objective function for $D$ provides lower bound of the dual cost function for the Wasserstein distance between $p$ and $p_G$.
\item Suggesting approximated algorithms and verifying that they improve Earth Mover's distance (EMD) \cite{flamary2017pot}, inception score \cite{salimans2016improved} and Fr\'echet inception distance (FID) \cite{heusel2017gans}.
\item Pointing out a \textit{generality} on DOT, i.e. if the $G$'s output domain is same as the $D$'s input domain, then we can use \textit{any} pair of trained $G$ and $D$ to improve generated samples.
\end{itemize}
In addition, we show some results on experiment comparing DOT and a naive method of
 improving sample just by increasing the value of $D$, under a fair setting.
One can download our codes from {\tt \href{https://github.com/AkinoriTanaka-phys/DOT}{https://github.com/AkinoriTanaka-phys/DOT}}.

\begin{figure}[t]
\centering
%
\includegraphics[width=14cm]{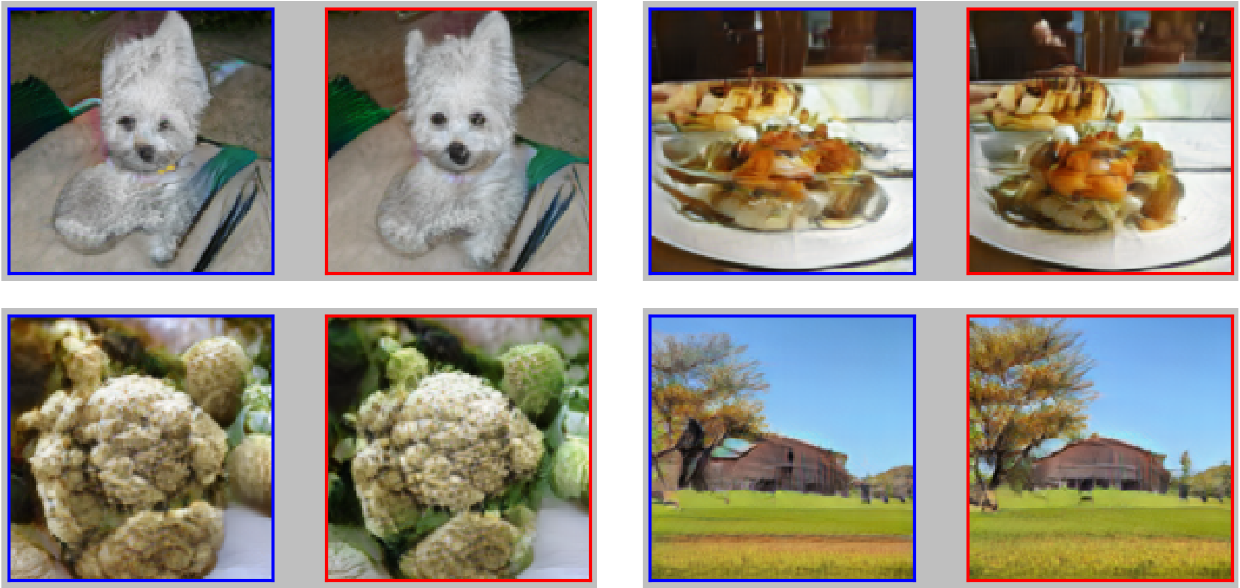}
\vspace*{-2mm}
\caption{
Each left image (blue): a sample from generator $G$.
Each right image (red): the sample modified by our algorithm based on discriminator $D$.
We use here the trained model available on {\tt \href{https://github.com/pfnet-research/sngan_projection}{https://github.com/pfnet-research/sngan\_projection}}
.
}
\label{fig:ot_all_imagenet}
\end{figure}

\section{Background}
\subsection{Generative Adversarial Nets}
Throughout this paper, we regard an image sample as a vector in certain Euclidean space:
$\bm x \in X$.
We name latent space as $Z$ and a prior distribution on it as $p_Z(\bm z)$.
The ultimate goal of the GAN is making generator $G : Z \to X$ whose push-foward of the prior
$
p_G(\bm x)
=
\int_Z p_Z(\bm z) \delta \big(\bm x - G(\bm z) \big) d \bm z
$
reproduces data-generating probability density $p(\bm x)$.
To achieve it, discriminator $D : X \to \mathbb{R}$ and
objective functions,
\begin{align}
&V_D(G, D)
=
\E{\sbm x\sim p}{f (-D(\bm x))}
+\E{\sbm y \sim p_G}{f(D(\bm y))}
,
\label{VD}
\\
&V_G(G, D)
=
\E{\sbm y \sim p_G}{g(D(\bm y))} 
,
\label{VG}
\end{align}
are introduced.
Choice of functions $f$ and $g$ corresponds to choice of GAN update algorithm as explained below.
Practically, $G$ and $D$ are parametric models $G_\theta$ and $D_\varphi$, and they are alternatingly updated as
\begin{align}
&\varphi \leftarrow \varphi + \epsilon  \nabla_\varphi V_D(G_\theta, D_\varphi),
\label{18}
\\
&\theta \leftarrow \theta - \epsilon  \nabla_\theta V_G(G_\theta, D_\varphi),
\label{17}
\end{align}
until the updating dynamics reaches to an equilibrium.
One of well know choices for $f$ and $g$ is
\begin{align}
f(u) = - \log (1+e^u) 
\quad
g(u) = - f(-u).
\label{gan}
\end{align}
Theoretically speaking, it seems better to take $g(u) = f(u)$, which is called minimax GAN \cite{fedus2017many} to guarantee $p_G = p$ at the equilibrium as proved in \cite{goodfellow2014generative}.
However, it is well known that taking \eqref{gan}, called non-saturating GAN, enjoys better performance practically.
As an alternative, we can choose the following $f$ and $g$ \cite{lim2017geometric, zhao2016energy}:
\begin{align}
f(u) = \max(0, -1-u),
\quad
g(u) = -u.
\label{hinge}
\end{align}
It is also known to be relatively stable.
In addition to it, $p_G=p$ at an equilibrium is proved at least in the theoretically ideal situation.
Another famous choice is taking
\begin{align}
f(u) = -u, 
\quad
g(u) = u.
\label{wgan}
\end{align}
The resultant GAN is called WGAN \cite{arjovsky2017wasserstein}.
We use \eqref{wgan} with gradient penalty (WGAN-GP) \cite{gulrajani2017improved} in our experiment.
WGAN is related to the concept of the optimal transport (OT) which we review below, so one might think our method is available only when we use WGAN.
But we would like to emphasize that such OT approach is also useful even when we take GANs described by \eqref{gan} and \eqref{hinge} as we will show later.
\subsection{Spectral normalization}
Spectral normalization (SN) \cite{miyato2018spectral} is one of standard normalizations on neural network weights to stabilize training process of GANs.
To explain it, let us define a norm for function called Lipschitz norm,
\begin{align}
||f||_{Lip}
:=
\sup \Big\{ 
\frac{ || f(\bm x) - f(\bm y) ||_2}{||\bm x-\bm y||_2} 
\Big|
\bm x \neq \bm y
\Big\}.
\label{lip}
\end{align}
For example,
$
||ReLU||_{Lip}
=
||lReLU||_{Lip}
=1
$
because their maximum gradient is 1.
For linear transformation $l_{W, b}$ with weight matrix $W$ and bias $b$, the norm $||l_{W, b}||_{Lip}$ is equal to the maximum singular value $\sigma(W)$.
Spectral normalization on $l_{W, b}$ is defined by dividing the weight $W$ in the linear transform by the $\sigma(W)$:
\begin{align}
SN(l_{W, b} )
=
l_{W/\sigma(W), b}.
\end{align}
By definition, it enjoys $||l_{W/\sigma(W)}||_{Lip} = 1$.
If we focus on neural networks, estimation of the upper bound of the norm is relatively easy because of the following property\footnote{
This inequality can be understood as follows.
Naively, the norm \eqref{lip} is defined by the maximum gradient between two different points.
Suppose $\bm x_1$ and $\bm x_2$ realizing maximum of gradient for $g$ and $\bm u_1$ and $\bm u_2$ are points for $f$, then the (RHS) of the inequality \eqref{ineq} is equal to $|| f(\bm u_1) - f(\bm u_2) ||_2/ || \bm u_1 - \bm u_2 ||_2 \times || g(\bm x_1) - g(\bm x_2) ||_2/ || \bm x_1 - \bm x_2 ||_2$.
If $g(\bm x_i) = \bm u_i$, it reduces to the (LHS) of the \eqref{ineq}, but the condition is not satisfied in general, and the (RHS) takes a larger value than (LHS).
This observation is actually important to the later part of this paper because estimation of the norm based on the inequality seems to be overestimated in many cases. 
}:
\begin{align}
||f \circ g ||_{Lip}
\leq ||f||_{Lip} \cdot || g ||_{Lip}
.
\label{ineq}
\end{align}
For example, suppose $f_{nn}$ is a neural network with ReLU or lReLU activations and spectral normalizations:
$
f_{nn}(\bm x)
=
SN \circ l_{W_L} \circ f  \circ SN \circ l_{W_{L-1}} \circ \dots  \circ SN \circ l_{W_1} (\bm x)
$, then the Lipschitz norm is bounded by one:
\begin{align}
||f_{nn}||_{Lip}
\leq
\prod_{l=1}^L || l_{W_l/\sigma(W_l)} ||_{Lip}
=1
\label{fnn1}
\end{align}
Thanks to this Lipschitz nature, the normalized network gradient behaves mild during repeating updates \eqref{18} and \eqref{17}, and as a result, it stabilizes the wild and dynamic optimization process of GANs.


\subsection{Optimal transport}
Another important background in this paper is optimal transport.
Suppose there are two probability densities, $p(\bm x)$ and $q(\bm y)$ where $\bm x, \bm y \in X$.
Let us consider the cost for transporting one unit of mass from $\bm x \sim p$ to $\bm y \sim q$.
The optimal cost is called Wasserstein distance.
Throughout this paper, we focus on the Wasserstein distance defined by $l_2$-norm cost $||\bm x - \bm y||_2$:
\begin{align}
W(p, q)
=
\min_{\pi \in \Pi(p, q)}
\Big(
\mathbb{E}_{(\sbm x,  \sbm y) \sim \pi}
\Big[
 ||\bm x-\bm y||_2 
 \Big]
\Big).
\label{eWD}
\end{align}
$\pi$ means joint probability for transportation between $\bm x$ and $\bm y$.
To realize it, we need to restrict $\pi$ satisfying marginality conditions,
\begin{align}
\int d\bm x \ \pi(\bm x, \bm y) = q(\bm y), \quad \int d\bm y \ \pi(\bm x, \bm y) = p(\bm x).
\label{pi_const}
\end{align}
An optimal $\pi^*$ satisfies $W(p, q) = \mathbb{E}_{(\sbm x, \sbm y) \sim \pi^*}[||\bm x - \bm y||_2]$, and it realizes the most effective transport between two probability densities under the $l_2$ cost.
Interestingly, \eqref{eWD} has the dual form
\begin{align}
W(p, q)
=
\max_{||\tilde{D}||_{Lip}  \leq 1}
\Big(
\mathbb{E}_{\sbm x\sim p}
\Big[
\tilde{D}(\bm x)
\Big]
-
\mathbb{E}_{\sbm y\sim q}
\Big[
\tilde{D}(\bm y)
\Big]
\Big).
\label{KantDual}
\end{align}
The duality is called Kantorovich-Rubinstein duality \cite{villani2008optimal, peyre2017computational}.
Note that $||f||_{Lip}$ is defined in \eqref{lip}, and the dual variable $\tilde{D}$ should satisfy Lipschitz continuity condition $||\tilde{D}||_{Lip} \leq 1$.
One may wonder whether any relationship between the optimal transport plan $\pi^*$ and the optimal dual variable $D^*$ exist or not.
The following theorem is an answer and it plays a key role in this paper.
%
\begin{theorem}
\label{th:1}
Suppose $\pi^*$ and $D^*$ are optimal solutions of the primal \eqref{eWD} and the dual \eqref{KantDual} problem, respectively. If $\pi^*$ is deterministic optimal transport described by a certain automorphism\footnote{
It is equivalent to assume there exists a solution of the corresponding Monge problem:
\begin{align}
\min_{T:X \to X} \Big(
\mathbb{E}_{\sbm y \sim q} \Big[
|| T(\bm y) - \bm y ||_2
\Big]
\Big),
\quad
\text{constrained by \eqref{reprod}.}
\notag
\end{align}
Reconstructing it from $\pi^*$ without any assumption is subtle problem and only guaranteed within strictly convex cost functions \cite{gangbo1996geometry}.
Although it is not satisfied in our $l_2$ cost,
there is a known method \cite{caffarelli2002constructing} to find a solution based on relaxing the cost to strictly convex cost $||\bm x-\bm y||^{1+\epsilon}_2$ with $\epsilon > 0$.
In our experiments, DOT works only when $||\bm x-\bm y||_2$ is small enough for given $\bm y$,
and it may suggest DOT approximates their solution, however, note that it is not evident whether our practical gradient-based implementation realizes it as pointed out in \cite{song2020discriminator}. 
} $T: X \to X$, then the following equations are satisfied:
\begin{align}
&||D^*||_{Lip} = 1,
\label{15lip}
\\
&T(\bm y)
\in
\argmin_{\sbm x} \Big\{
||\bm x - \bm y ||_2
- D^*(\bm x)
\Big\}
,
\label{CondEn}
\\
&
p(\bm x)
=
\int d \bm y \ \delta \Big( \bm x - T(\bm y) \Big) q(\bm y)
.
\label{reprod}
\end{align}
\end{theorem}
(Proof)
It can be proved by combining well know facts.
See Supplementary Materials. $_\square$
%
\section{Discriminator optimal transport}
If we apply the spectral normalization on a discriminator $D$, it satisfies K-Lipschitz condition $||D||_L = K$ with a certain real number $K$.
By redefining it to $\tilde{D} = D/K$, it becomes 1-Lipschitz $||\tilde{D}||_L = 1$.
It reminds us the equation \eqref{15lip}, and one may expect a connection between OT and GAN.
In fact, we can show the following theorem:
\begin{theorem}
\label{th:2}
Each objective function of GAN using logistic \eqref{gan}, or hinge \eqref{hinge}, or identity \eqref{wgan} loss with gradient penalty, provides lower bound of the mean discrepancy of $\tilde{D} = D/K$ between $p$ and $p_G$:
\begin{align}
V_D(G, D)
\leq
K \Big(
\mathbb{E}_{\sbm x\sim p}
\Big[
\tilde{D}(\bm x)
\Big]
-
\mathbb{E}_{\sbm y\sim p_G}
\Big[
\tilde{D}(\bm y)
\Big]
\Big).
\label{key2}
\end{align}
\end{theorem}
(Proof)
See Supplementary Materials. $_\square$

In practical optimization process of GAN, $K$ could change its value during the training process, but it stays almost constant at least approximately as explained below.
\subsection{Discriminator Optimal Transport (ideal version)}
The inequality \eqref{key2} implies that the update \eqref{18} of $D$ during GANs training maximizes the lower bound of the objective in \eqref{KantDual}, the dual form of the Wasserstein distance.
In this sense, the optimization of $D$ in \eqref{18} can be regarded solving the problem \eqref{KantDual} approximately\footnote{
This situation is similar to guarantee VAE \cite{kingma2013auto} objective function which is a lower bound of the likelihood called evidence lower bound (ELBO). Note that however, our inequalities are strictly bounded except for \eqref{hinge}.
}.
If we apply \eqref{CondEn} with $D^* \approx \tilde{D} = D/K$,
the following transport of given $\bm y \sim p_G$
\begin{align}
T_D(\bm y) \in \argmin_{\sbm x} \Big\{
||\bm x- \bm y||_2
- \frac{1}{K} D(\bm x)
\Big\}
\end{align}
is expected to improve the sample because of the Theorem \ref{th:1}.
\subsection{Discriminator Optimal Transport (practical version)}
To check whether $K$ changes drastically or not during the GAN updates, we calculate approximated Lipschitz constants defined by
\begin{align}
&K_\text{eff}
=
\max \Big\{
\frac{|D(\bm x) - D(\bm y)|}{||\bm x- \bm y||_2}
\Big|
\bm x , \bm y \sim p_G 
\Big\},
\label{Keff}
\\
&k_\text{eff}
=
\max \Big\{
\frac{| D \circ G(\bm z) -  D \circ G(\bm z_{\sbm y})|}{||\bm z-\bm z_{\sbm y}||_2}
\Big|
\bm z , \bm z_{\sbm y} \sim p_Z 
\Big\},
\label{keff}
\end{align}
in each 5,000 iteration on GAN training with CIFAR-10 data with DCGAN models explained in Supplementary Materials.
As plotted in Figure \ref{fig:ks}, both of them do not increase drastically.
\begin{figure}[t]
\centering
\includegraphics[width=400pt]{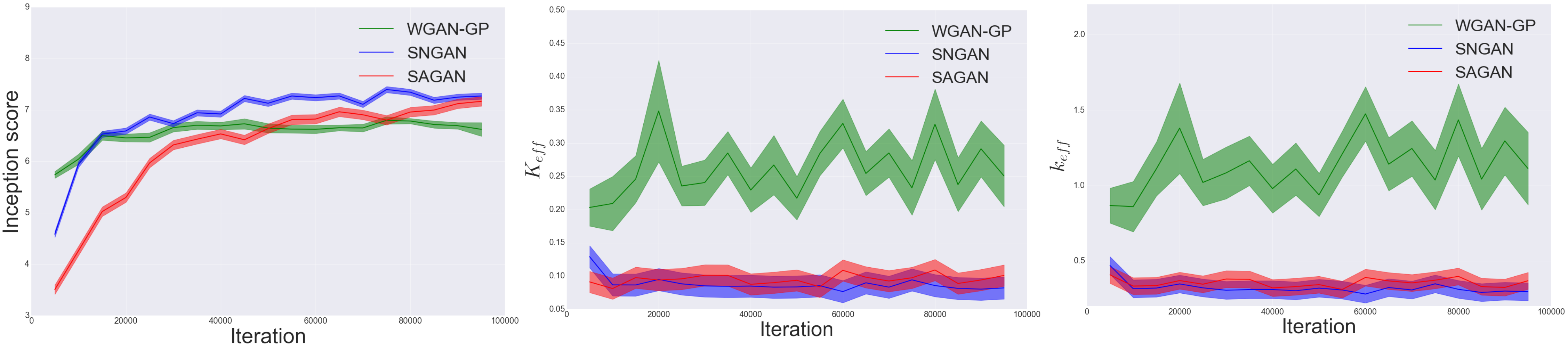}
\vspace*{-2mm}
\caption{
Logs of inception score (left), approximated Lipschitz constant of $D$ (middle), and approximated Lipschitz constant of $D \circ G$ (right) on each GAN trained with CIFAR-10. Approximated Lipschitz constants are calculated by random 500 pair samples.
Errorbars are plotted within 1$\sigma$ by 500 trials.
}
\label{fig:ks}
\end{figure}
It is worth to mention that the naive upper bound of the Lipschitz constant like \eqref{fnn1} turn to be overestimated. For example, SNGAN has the naive upper bound 1,  but \eqref{Keff} stays around 0.08 in Figure \ref{fig:ks}.

\paragraph{Target space DOT}
Based on these facts, we conclude that trained discriminators can approximate the optimal transport \eqref{CondEn} by 
\begin{align}
T_D^\text{eff}(\bm y)
\in
\argmin_{\sbm x} \Big\{ || \bm x- \bm y||_2 -  \frac{1}{K_\text{eff}} D(\bm x) \Big\}.
\label{DOT}
\end{align}
As a preliminary experiment, we apply DOT to WGAN-GP trained by 25gaussians dataset and swissroll dataset.
We use the gradient descent method shown in Algorithm \ref{alg1} to search transported point $T_D^\text{eff}(\bm y)$ for given $\bm y \sim p_G$.
In Figure \ref{2d_WGANGP}, we compare the DOT samples and naively transported samples by the discriminator which is implemented by replacing the gradient in Algorithm \ref{alg1} to
$
- \frac{1}{K_\text{eff}} \nabla_{\sbm x} D(\bm x) 
$
, i.e. just searching $\bm x$ with large $D(\bm x)$ from initial condition $\bm x \leftarrow \bm y$ where $\bm y \sim p_G$.
\begin{algorithm}[t]                      
\caption{Target space optimal transport by gradient descent}         
\label{alg1}                          
\begin{algorithmic}                  
\REQUIRE trained $D$,  approximated $K_\text{eff}$ by \eqref{Keff}, sample $\bm y$, learning rate $\epsilon$ and small vector $\bm \delta$
\STATE Initialize $\bm x \leftarrow \bm y$
\FOR{$n_\text{trial}$ in range($N_\text{updates}$)}
\STATE {$\bm x \leftarrow \bm x - \epsilon \bm \nabla_{\sbm x} \Big\{ ||\bm x - \bm y + \bm \delta||_2 - \frac{1}{K_\text{eff}} D(\bm x) \Big\}$
\quad
(
\text{$\bm \delta$ is for preventing overflow.}
)
}
\ENDFOR
\RETURN{$\bm x$}
\end{algorithmic}
\end{algorithm}

DOT outperforms the naive method qualitatively and quantitatively.
On the 25gaussians, one might think 4th naively improved samples are better than 3rd DOT samples.
However, the 4th samples are too concentrated and lack the variance around each peak.
In fact, the value of the Earth Mover's distance, EMD, which measures how long it is separated from the real samples, shows relatively large value.
On the swissroll, 4th samples based on naive transport lack many relevant points close to the original data, and it is trivially bad. On the other hand, one can see that the 3rd DOT samples keep swissroll shape and clean the blurred shape in the original samples by generator.

\begin{figure}[t]
\centering
\includegraphics[width=400pt]{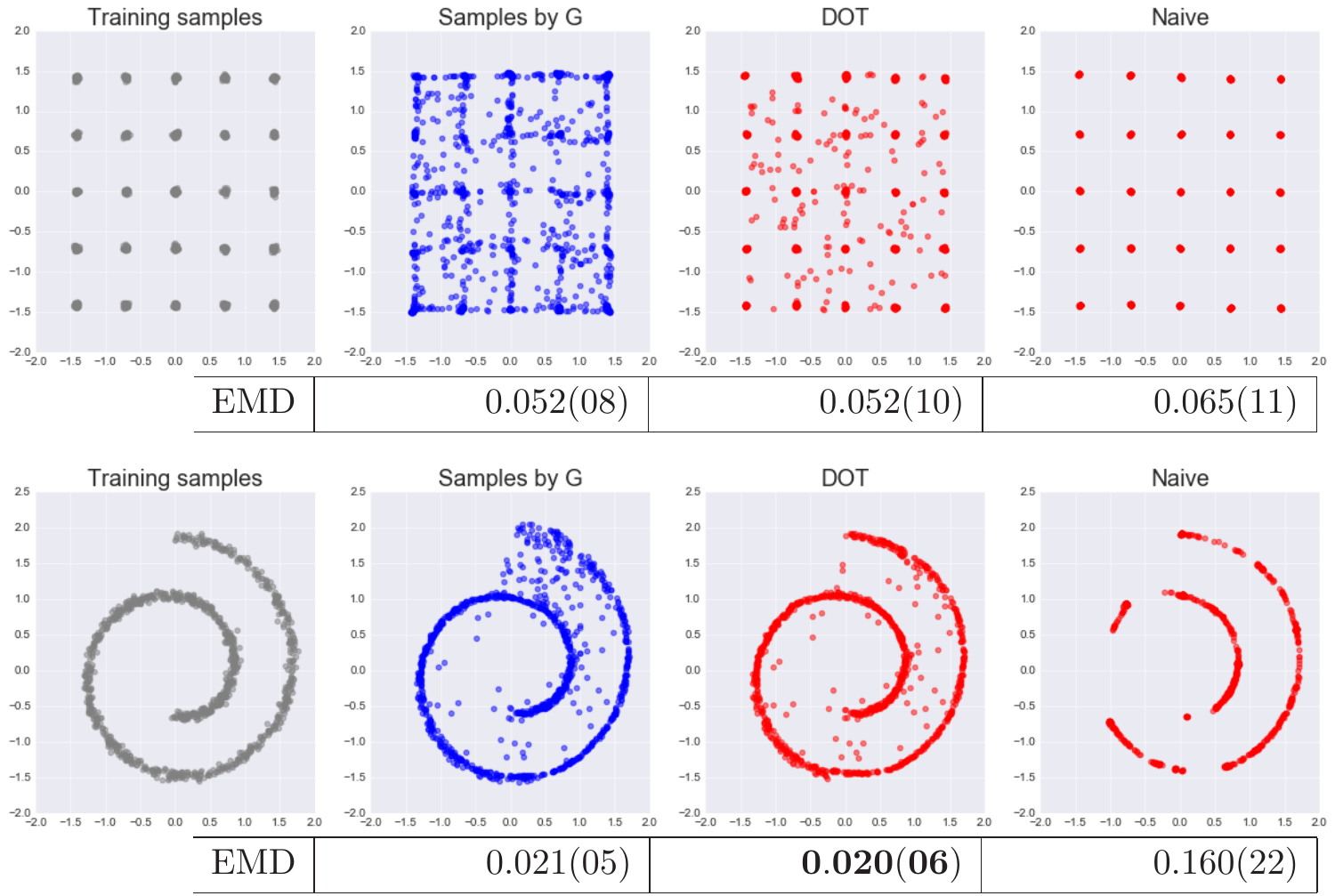}
\caption{
2d experiments by using trained model of WGAN-GP. 
1,000 samples of, 1st: training samples,
2nd: generated samples by $G$,
3rd: samples by target space DOT,
4th: samples by naive transport, are plotted.
Each EMD value is calculated by 100 trials.
The error corresponds to 1$\sigma$.
We use $\bm \delta = {\bf 0.001}$.
See the supplementary material for more details on this experiment.
}
\label{2d_WGANGP}
\end{figure}
\paragraph{Latent space DOT}
The target space DOT works in low dimensional data, but it turns out to be useless once we apply it to higher dimensional data. See Figure \ref{fig:dot2} for example.
Alternative, and more workable idea is regarding $D \circ G : Z \to \mathbb{R}$ as the dual variable for defining Wasserstein distance between ``pullback'' of $p$ by $G$ and prior $p_Z$.
Latent space OT itself is not a novel idea \cite{agustsson2017optimal, salimans2018improving}, but there seems to be no literature using trained $G$ and $D$, to the best of our knowledge.
The approximated Lipschitz constant of $G \circ D$ also stays constant as shown in the right sub-figure in Figure \ref{fig:ks}, so we propose
\begin{align}
T_{D \circ G}^\text{eff}(\bm z_{\sbm y})
\in
\argmin_{\sbm z} \Big\{
|| \bm z - \bm z_{\sbm y} ||_2 - \frac{1}{k_\text{eff}} D \circ G(\bm z)
\Big\} 
\end{align}
to approximate optimal transport in latent space.
Note that if the prior $p_Z$ has non-trivial support, we need to restrict $\bm z$ onto the support during the DOT process.
In our algorithm \ref{alg2}, we apply projection of the gradient.
One of the major practical priors is normal distribution $\mathcal{N}(0, \bm I_{D\times D})$ where $D$ is the latent space dimension.
If $D$ is large, it is well known that the support is concentrated on $(D-1)$-dimensional sphere with radius $\sqrt{D}$, so the projection of the gradient $\bm g$ can be calculated by $\bm g - (\bm g \cdot \bm z) \bm z/ \sqrt{D} $ approximately.
Even if we skip this procedure, transported images may look improved, but it downgrades inception scores and FIDs. 

\begin{algorithm}[t]                      
\caption{Latent space optimal transport by gradient descent}         
\label{alg2}                          
\begin{algorithmic}                  
\REQUIRE trained $G$ and $D$,  approximated $k_\text{eff}$, sample $\bm z_{\sbm y}$, learning rate $\epsilon$, and small vector $\bm \delta$
\STATE Initialize $\bm z \leftarrow \bm z_{\sbm y}$
\FOR{$n_\text{trial}$ in range($N_\text{updates}$)}
\STATE {$\bm g = \bm \nabla_{\sbm z} \Big\{ ||\bm z - \bm z_{\sbm y} + \bm \delta||_2 - \frac{1}{k_\text{eff}} D\circ G(\bm z) \Big\}$
\quad
(
\text{$\bm \delta$ is for preventing overflow.}
)
}
\IF {noise is generated by $\mathcal{N}(0, \bm I_{D\times D})$}
\STATE {$\bm g \leftarrow \bm g - (\bm g \cdot \bm z) \bm z / \sqrt{D} $}
\ENDIF
\STATE {$\bm z \leftarrow \bm z - \epsilon \bm g$}
\IF {noise is generated by $U([-1, 1])$}
\STATE {clip $\bm z \in [-1, 1]$ }
\ENDIF
\ENDFOR
\RETURN{$\bm x = G(\bm z)$}
\end{algorithmic}
\end{algorithm}


\section{Experiments on latent space DOT}
\begin{figure}[t]
\centering
\includegraphics[width=150pt]{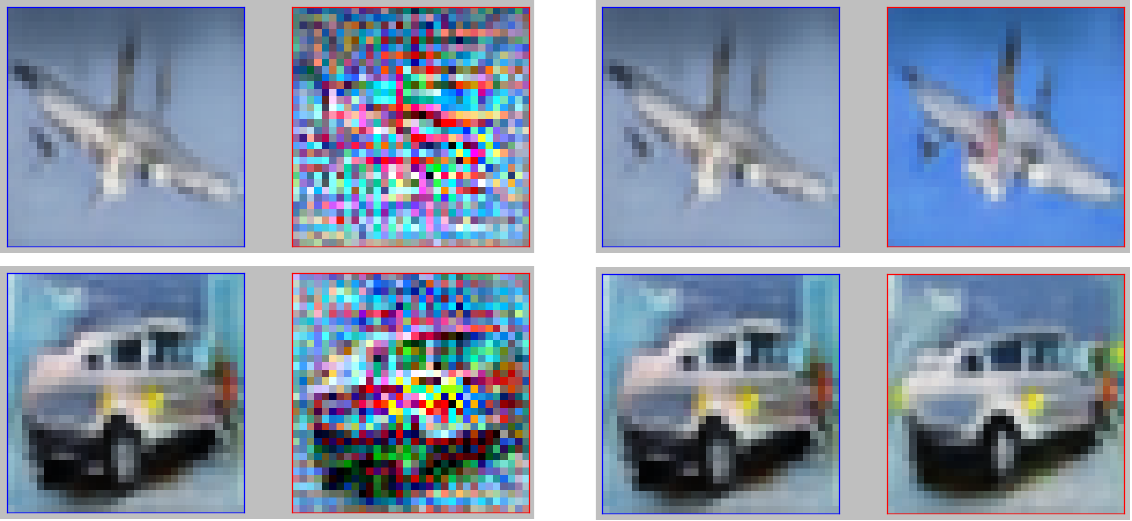}
\quad
\includegraphics[width=150pt]{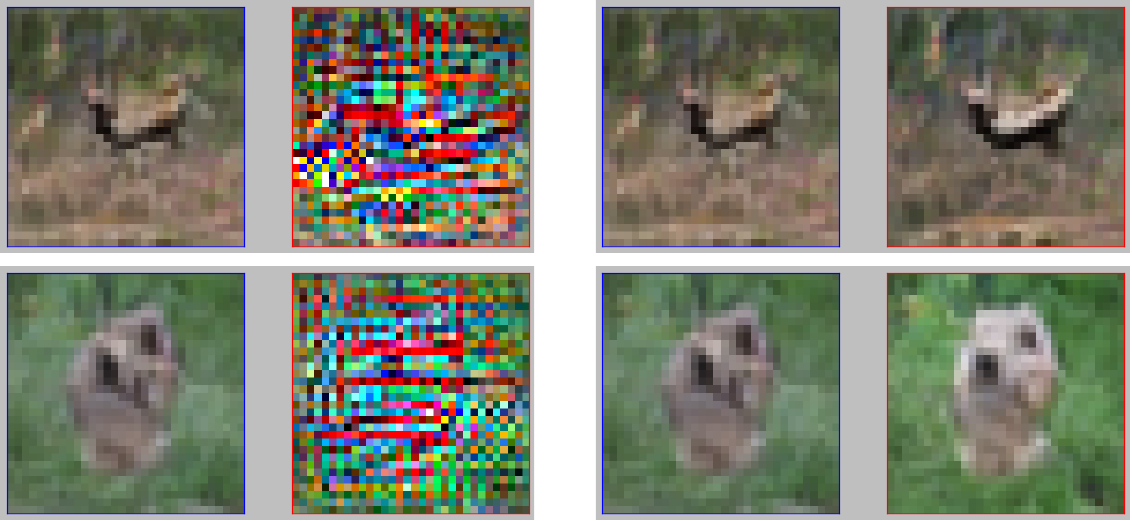}
\vspace*{-2mm}
\caption{
Target space DOT sample (each left) and latent space DOT sample (each right).
The former looks giving meaningless noises like perturbations in adversarial examples \cite{szegedy2013intriguing}.
On the other hand, the latent space DOT samples keep the shape of image, and clean it.
}
\label{fig:dot2}
\end{figure}

\subsection{CIFAR-10 and SLT-10}\label{sec41}
%
We prepare pre-trained DCGAN models and ResNet models on various settings, and apply latent space DOT. 
In each case, inception score and FID are improved (Table \ref{tab:scores}).
We can use arbitrary discriminator $D$ to improve scores by fixed $G$ as shown in Table \ref{tab:intriguing}.
As one can see, DOT really works.
But it needs tuning of hyperparameters.
First, it is recommended to use small $\epsilon$ as possible.
A large $\epsilon$ may accelerate upgrading, but easily downgrade unless appropriate $N_\text{updates}$ is chosen.
Second, we recommend to use $k_\text{eff}$ calculated by using enough number of samples.
If not, it becomes relatively small and it also possibly downgrade images.
As a shortcut, $k_\text{eff}=1$ also works.
See Supplementary Materials for details and additional results including comparison to other methods.
\begin{table}[b]
\vspace{-.6cm}
\centering
  \begin{tabular}{p{9mm}p{17mm}|cc|cc|}
 &
 & \multicolumn{2}{|c|}{CIFAR-10}
 & \multicolumn{2}{|c|}{STL-10}
 \\ \hline
   &  & bare & DOT 
      & bare & DOT \\
 \hline
{\small DCGAN} & WGAN-GP
& 6.53(08), 27.84
& 7.45(05), 24.14
& 8.69(07), 49.94
& 9.31(07), 44.45
\\ 
& SNGAN(ns)
& 7.45(09), 20.74
& 7.97(14), {\bf 15.78}
& 8.67(01), 41.18
& 9.45(13), {\bf 34.84}
\\ 
& SNGAN(hi)
& 7.45(08), 20.47
& 8.02(16), 17.12
& 8.83(12), 40.10
& 9.35(12), 34.85
\\ 
& SAGAN(ns)
& 7.75(07), 25.37
& {\bf 8.50(01)}, 20.57
& 8.68(01), 48.23
& 10.04(14), 41.19
\\ 
& SAGAN(hi)
& 7.52(06), 25.78
& 8.38(05), 21.21
& 9.29(13), 45.79
& {\bf 10.30(21)}, 40.51
\\ \hline
{\small Resnet} & SAGAN(ns)
& 7.74(09), 22.13
& 8.49(13), 20.22
& 9.33(08), 41.91
& {\bf 10.03(14), 39.48}
\\
& SAGAN(hi)
& 7.85(11), 21.53
& {\bf 8.50(12), 19.71}
& 
& 
\\ \hline
  \end{tabular}
 \caption{(Inception score, FID) by usual sampling (bare) and DOT:
Models in \cite{miyato2018spectral} and self-attention layer \cite{zhang2018self} are used.
(ns) and (hi) mean models trained by \eqref{gan} and \eqref{hinge}.
$\epsilon=0.01$ SGD is applied 20 times for CIDAR-10 and 10 times for STL-10.
$k_\text{eff}$ is calculated by 100 samples and $\bm \delta = {\bf 0.001}$.
 }
 \label{tab:scores}
\
 \begin{tabular}{l|c||c|c|c|c|c|}
$D$& 
without $D$ &
 WGAN-gp &
 SNGAN(ns) &
 SNGAN(hi) &
 SAGAN(ns) &
 SAGAN(hi) \\
 \hline
IS & 
7.52(06)
&
8.03(11)
&
8.22(07)
&
8.38(07)
&
8.36(12)
&
8.38(05)
\\ \hline 
FID&
25.78
&
24.47
&
21.45
&
23.03
&
21.07
&
21.21
\\ \hline
 \end{tabular}
 \caption{Results on scores by $G_\text{SAGAN(ns)}$ after latent space DOT using each $D$ in different training scheme using CIFAR-10 within DCGAN architecture.
Parameters for DOT are same in Table \ref{tab:scores}.
 }
 \label{tab:intriguing}
\end{table}

\subsection{ImageNet}
\begin{figure}[t]
\centering
\includegraphics[width=400pt]{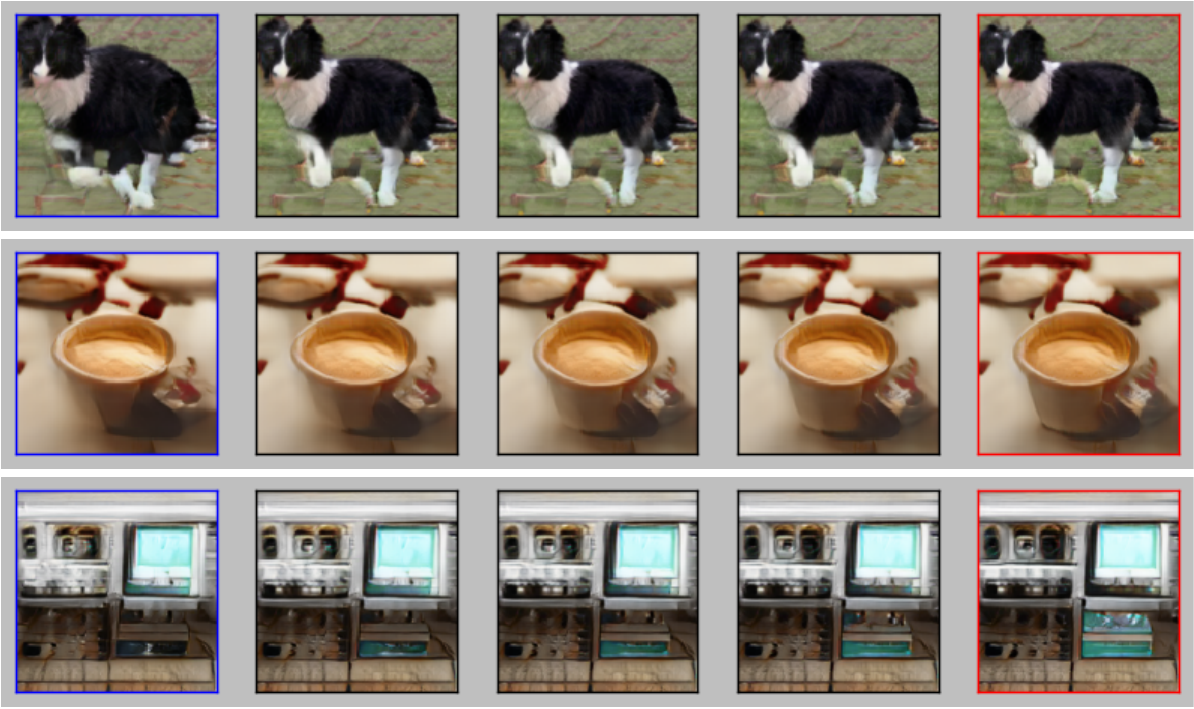}
\caption{
Left images surrounded by blue lines are samples from the conditional generator.
The number of updates $N_\text{updates}$ for DOT increases along horizontal axis.
Right Images surrounded by red lines corresponds after 30 times updates with Adam $(\alpha, \beta_1, \beta_2) = (0.01, 0, 0.9)$ and $k_\text{eff}(y) = 1$.
}
\label{fig:cDOT}
\end{figure}
\paragraph{Conditional version of latent space DOT}
In this section, we show results on ImageNet dataset.
As pre-trained models,  we utilize a pair of public models $(G, D)$ \cite{miyato2018cgans} of conditional GAN \cite{mirza2014conditional} (available at {\tt \href{https://github.com/pfnet-research/sngan_projection}{https://github.com/pfnet-research/sngan\_projection}}.)
In conditional GAN, $G$ and $D$ are networks conditioned by label $y$.
Typical objective function $V_D$ is therefore represented by average over the label:
\begin{align}
V_D(G, D)
=
\mathbb{E}_{y \sim p(y)}
\Big[
V_D\Big(G(\cdot| y) , D(\cdot | y) \Big)
\Big]
.
\end{align}
But, once $y$ is fixed, $G(\bm z| y)$ and $D(\bm x| y)$ can be regarded as usual networks with input $\bm z$ and $\bm x$ respectively.
So, by repeating our argument so far, DOT in conditional GAN can be written by
\begin{align}
T_{G \circ D}(\bm z_{\scriptsize \bm y} | y)
\in
\text{argmin}_{ \sbm z }
\Big\{
||\bm z- \bm z_{\sbm y}||_2
-
\frac{1}{k_\text{eff} (y)}
D \Big( G(\bm z | y)  \big| y \Big)
\Big\}.
\label{cDOT}
\end{align}
where $k_\text{eff} (y)$ is approximated Lipschitz constant conditioned by $y$.
It is calculated by
\begin{align}
&k_\text{eff}(y)
=
\max \Big\{
\frac{| D \big( G(\bm z|y) \big| y \big) -  D \big( G(\bm z_{\sbm y} | y) \big| y \big) |}{||\bm z-\bm z_{\sbm y}||_2}
\Big|
\bm z , \bm z_{\sbm y} \sim p_Z 
\Big\}.
\label{ckeff}
\end{align}
\paragraph{Experiments}
We apply gradient descent updates with with Adam$(\alpha, \beta_1, \beta_2)=(0.01, 0, 0.9)$.
\if
Inception score and FID are improved as
\begin{align}
36.65(54), 43.35
\quad
\to
\quad
\bm{37.34(69), 42.47}
.
\end{align}
As one can see, DOT improves these scores even with very naive estimation of the Lipschitz constant.
\fi
%
We show results on 4 independent trials in Table \ref{tab:imagenet}.
It is clear that DOT mildly improve each score.
Note that we need some tunings on hyperparameters $\epsilon, N_\text{updates}$ as we already commented in \ref{sec41}.
%
\begin{table}[b]
\vspace{-1cm}
\centering
 \begin{tabular}{l||c|c|c|c|}
 &
\text{\# updates}=0 
& \text{\# updates}=4 
& \text{\# updates}=16 
& \text{\# updates}=32
 \\ \hline
 \text{trial1($k_\text{eff} (y) = 1$)} 
 &
36.40(91), 43.34 & 
36.99(75), 43.01 & 
37.25(84), 42.70 & 
{\bf 37.61(88), 42.35}  
\\
 \text{trial2($k_\text{eff} (y) = 1$)}  &
 36.68(59), 43.60 & 
36.26(98), 43.09 & 
36.97(63), 42.85 & 
{\bf 37.02(73), 42.74}  
\\
 \text{trial3} &
 36.64(63), 43.55 & 
36.87(84), 43.11 & 
{\bf 37.51(01), 42.43} & 
36.88(79), 42.52  
\\
 \text{trial4} &
36.23(98), 43.63 & 
36.49(54), 43.25 & 
37.29(86), 42.67 & 
{\bf 37.29(07), 42.40}  
  \end{tabular}
 \caption{
(Inception score, FID) for each update.
Upper 2 cases are executed by $k_\text{eff}(y)=1$ without calculating \eqref{ckeff}.
We use 50 samples for each label $y$ to calculate $k_\text{eff}(y)$ in lower 2 trials.
$\bm \delta = {\bf 0.001}$.
 }
 \label{tab:imagenet}
\end{table}

\newpage
\begin{algorithm}[h]                      
\caption{Latent space conditional optimal transport by gradient descent}         
\label{alg3}                          
\begin{algorithmic}                  
\REQUIRE trained $G$ and $D$,  label $y$, approximated $k_\text{eff}(y)$, sample $\bm z_{\sbm y}$, learning rate $\epsilon$ and small vector $\bm \delta$
\STATE Initialize $\bm z \leftarrow \bm z_{\sbm y}$
\FOR{$n_\text{trial}$ in range($N_\text{update}$)}
\STATE {$\bm g = \bm \nabla_{\sbm z} \Big\{ ||\bm z - \bm z_{\sbm y} + \bm \delta||_2 - \frac{1}{k_\text{eff}(y)} D \Big( G(\bm z | y) \Big| y \Big) \Big\}$
\quad
(
\text{$\bm \delta$ is for preventing overflow.}
)
}
\IF {noise is generated by $\mathcal{N}(0, \bm I_{D\times D})$}
\STATE {$\bm g \leftarrow \bm g - (\bm g \cdot \bm z) \bm z / \sqrt{D} $}
\ENDIF
\STATE {$\bm z \leftarrow \bm z - \epsilon \bm g$}
\IF {noise is generated by $U([-1, 1])$}
\STATE {clip $\bm z \in [-1, 1]$ }
\ENDIF
\ENDFOR
\RETURN{$\bm x = G(\bm z | y)$}
\end{algorithmic}
\end{algorithm}
\paragraph{Evaluation}
To calculate FID, we use available 798,900 image files in ILSVRC2012 dataset.
We reshape each image to the size $299 \times 299 \times 3$, feed all images to the public inception model to get the mean vector $\bm m_w$ and the covariance matrix $\bm C_w$ in 2,048 dimensional feature space.
%

\section{Conclusion}
In this paper, we show the relevance of discriminator optimal transport (DOT) method on various trained GAN models to improve generated samples.
Let us conclude with some comments here.

First, DOT objective function in \eqref{DOT} reminds us the objective for making adversarial examples \cite{szegedy2013intriguing}.
There is known fast algorithm to make adversarial example making use of the piecewise-linear structure of the ReLU neural network \cite{goodfellow2014explaining}.
The method would be also useful for accelerating DOT.

Second, latent space DOT can be regarded as improving the prior $p_Z$. A
similar idea can be found also in \cite{brock2018large}.
In the usual context of the GAN, we fix the prior, but it may be possible to train the prior itself simultaneously by making use of the DOT techniques.

\subsubsection*{Note added}
Although our proposed gradient-based algorithms work well, it is not evident whether the gradient update can find the exact transport map $T$. Gradient descent of the objective function on the right hand side of \eqref{CondEn} is expected to provide a minimum of the objective, that includes $T({\bm y})$, if it exists, but many other points represented by $\{t{\bm y} + (1-t)T({\bm y}) | t \in [0,1]  \}$ are also included in the set of minima as pointed out in \cite{song2020discriminator}. If we could remove these unnecessary points, it would be possible to make better update.

We leave these as future works.

\subsubsection*{Acknowledgments}
We would like to thank Asuka Takatsu for fruitful discussion and Kenichi Bannai for careful reading this manuscript.
This work was supported by computational resources provided by 
RIKEN AIP deep learning environment (RAIDEN) and RIKEN iTHEMS.



\input{endbib}
\appendix
\documentclass{article}

\newcommand{\inputt}[1]{\input{#1}}
\newcommand{\usefinal}{\usepackage[final]{neurips_2019}}
\input{begin_sup}

\newcommand{\gantable}[7]{
\begin{table}[th]
 \begin{minipage}{#1\hsize}
 \centering
 \begin{tabular}{c}
 \toprule
 \hline
 #2
\\
\hline
\bottomrule
{\small (i) Generator}
 \end{tabular}
\end{minipage}
 \begin{minipage}{#3\hsize}
\
\end{minipage}
 \begin{minipage}{#4\hsize}
 \centering
  \begin{tabular}{c}
 \toprule \hline
 #5
\\
\hline
\bottomrule
\small (ii) Discriminator
 \end{tabular}
\end{minipage}
\vspace{10pt}
 \caption{#6}
 \label{tab:#7}
\end{table}
}

\section{Proofs}

\subsection{Proof of Theorem 1}\label{Thm1}
We show a proof of {\bf Theorem 1} here by utilizing well known propositions in optimal transport \cite{villani2008optimal, peyre2017computational}.
First, we show the following proposition for later use.
\begin{proposition}
\textit{Suppose $\pi^*$ and $D^*$ are optimal solutions of primal and dual  problem respectively, then the equation}
\begin{align}
\int d\bm x d\bm y \ \pi^*(\bm x, \bm y)
\Big[
||\bm x- \bm y||_2
-
\Big(
D^*(\bm x)
-
D^*(\bm y)
\Big)
\Big]
=
0
\label{28}
\end{align}
\textit{is satisfied.}
\end{proposition}
(Proof)
Thanks to the strong duality, we have
\begin{align}
\mathbb{E}_{(\sbm x, \sbm y)\sim \pi^* }
\Big[
|| \bm x- \bm y||_2
\Big]
&=
\mathbb{E}_{\sbm x\sim p}
\Big[
D^*(\bm x)
\Big]
-
\mathbb{E}_{\sbm y\sim p_G}
\Big[
D^*(\bm y)
\Big].
\label{29}
\end{align}
Now, let us remind that $\pi^*(\bm x, \bm y)$ satisfies the marginality conditions
$p(\bm x) = \int d \bm y \ \pi^*(\bm x, \bm y)$ and $q(\bm y) = \int d \bm y \ \pi^*(\bm x, \bm y)$.
It means we can replace the (RHS) of \eqref{29} by
\begin{align}
\mathbb{E}_{(\sbm x, \sbm y)\sim \pi^*}
\Big[
D^*(\bm x)
-
D^*(\bm y)
\Big].
\end{align}
By transposing it to (LHS) of \eqref{29}, it completes the proof.$_\square$

As a corollary of the proposition, we can show the first identity in {\bf Theorem 1},
\begin{align}
&||D^*||_{Lip} = 
\sup \Big\{ \frac{|D^*(\bm x) - D^*(\bm y)|}{||\bm x - \bm y||_2} \Big| \bm x \neq \bm y \Big\}
=1.
\label{282}
\end{align}
First, let us remind that $||D^*||_{Lip} \leq 1$ is automatically satisfied.
It means for arbitrary $\bm x$ and $\bm y$, 
\begin{align}
\Big[
||\bm x- \bm y||_2
-
|
D^*(\bm x)
-
D^*(\bm y)
|
\Big]
\geq 0
\end{align}
is satisfied.
Next, 
$-x \geq -|x|$ is trivially true for arbitrary $x \in \mathbb{R}$.
By using this inequality with $x = D^*(\bm x)-D^*(\bm y)$, we conclude
\begin{align}
\Big[
||\bm x- \bm y||_2
-
\Big(
D^*(\bm x)
-
D^*(\bm y)
\Big)
\Big]
\geq0.
\end{align}
It means the integrand in the equation \eqref{28} is always positive or zero.
Then, we can say
\begin{align}
\pi^*(\bm x, \bm y)\neq0
\quad
\Rightarrow
\quad
\Big[
||\bm x- \bm y||_2
-
\Big(
D^*(\bm x)
-
D^*(\bm y)
\Big)
\Big]
=0,
\label{11}
\end{align}
because if not, we cannot cancel its contribution in the integral \eqref{28}.
$\pi^*$ is probability density, so there exists a pair $(\bm x, \bm y)$ satisfying $\pi^*(\bm x, \bm y)\neq0$, and the pair realizes the absolute gradient 1.
As already noted, $D^*$ should satisfy $||D^*||_{Lip}\leq 1$, so \eqref{11} means there exists two element $\bm x$ and $\bm y$ realizing this upper bound, i.e. $||D^*||_{Lip} = 1$.

The second equation 
\begin{align}
&T(\bm y) \in \argmin_{\sbm x} \Big\{ ||\bm x - \bm y||_2 - D^*(\bm x) \Big\}
,
\label{292}
\end{align}
is also proved as a corollary of {\bf Proposition 1}.
But we need to use a help of the assumption in {\bf Theorem 1}, i.e. the existence of the deterministic solution $T$ of the Monge's problem.
It means $\pi^*(\bm x, \bm y)$ is deterministic by a certain automorphism $T$ and
described by Dirac's delta function\footnote{
If we do not consider Wasserstein-1 but Wasserstein-$2$, there is no need to assume the existence of $T$ in advance and it is called Brenier's theorem \cite{villani2003topics}.
}  with respect to $\bm x$ for given $\bm y$, i.e.
\begin{align}
\left. \begin{array}{ll}
\pi^*(\bm x, \bm y) \neq 0 & \text{if } \bm x = T(\bm y), \\
\pi^*(\bm x, \bm y) = 0 & \text{otherwise.} \\
\end{array} \right.
\label{pistar}
\end{align}
Because of $||D^*||_{Lip} = 1$, 
\begin{align}
-D^*(\bm y)
\leq
||\bm x  - \bm y||_2 - D^*(\bm x )
\label{40}
\end{align}
is satisfied for arbitrary $\bm x$.
On the other hand, thanks to the equality \eqref{11},
\begin{align}
-D^*(\bm y)
=
||T(\bm y)- \bm y||_2 - D^*(T(\bm y))
\label{40-2}
\end{align}
should be satisfied. 
It means the (RHS) of \eqref{40-2} is the minimum value of (RHS) of \eqref{40}, and it completes the proof of \eqref{292}.

The third identity
\begin{align}
p(\bm x)
=
\int d \bm y \ \delta \Big( \bm x - T(\bm y) \Big) q(\bm y),
\label{37}
\end{align}
can be got as a corollary of the following proposition.

\begin{proposition}
If $\pi^*$ is the deterministic solution of the primal problem, then it should be represented by
\begin{align}
\pi^*(\bm x, \bm y) = \delta \Big( \bm x - T(\bm y) \Big) q(\bm y)
\end{align}
with the optimal transport map $T:X\to X$.
\end{proposition}
(Proof)
First of all, because of the assumption \eqref{pistar}, $\pi^*$ should be proportional to $\delta(\bm x - T(\bm y) )$.
To satisfy the marginal conditions of $\pi$, we multiply a function of $\bm x, \bm y$ to it:
\begin{align}
\pi^*(\bm x, \bm y)
=
\delta \Big(\bm x - T(\bm y) \Big)
f(\bm x, \bm y).
\end{align}
Thanks to the delta function, however, it is sufficient to take into account $f(T^{-1}(\bm y), \bm y)$ and let us call it $g(\bm y)$, then
\begin{align}
\pi^*(\bm x, \bm y)
=
\delta \Big( \bm x - T(\bm y) \Big) g(\bm y).
\end{align}
Now, let us consider the marginal condition on $\bm y$, i.e. integration over $\bm x$ should be equal to $q(\bm y)$:
\begin{align}
q(\bm y)
=
\int d\bm x \ \pi^*(\bm x, \bm y)
=
\int d\bm x \
\delta \Big( \bm x - T(\bm y) \Big) g(\bm y)
=
g(\bm y).
\end{align}
It completes the proof.$_\square$

\vspace{10pt}\noindent
In the above proof, we do not consider taking marginal along $\bm y$ which gives \eqref{37} by definition.
One may be suspicious on it.
In fact, by directly integrating it over $\bm y$, we get
\begin{align}
\int d\bm y \
\delta \Big( \bm x - T(\bm y) \Big) q(\bm y)
=
\frac{q(\bm y )}{
|\det \nabla_{\sbm y} T(\bm y)|
}
\Big|_{\sbm y = T^{-1} (\sbm x)}
.
\label{33}
\end{align}
But it is known that the (RHS) actually agree with $p(\bm x)$.
Physical meaning of this fact is simple. Now, let $T^{-1}: X \to X$ is the optimal transportation from $p(\bm x)$ to $q(\bm y)$.
The numerator of \eqref{33} is just a map of mass of the probability density, and the denominator corresponds to the Jacobian to guarantee its integration over $\bm x$ is 1.
\begin{align}
\int d \bm x
\frac{q(\bm y )}{
|\det \nabla_{\sbm y} T(\bm y)|
}
\Big|_{\sbm y = T^{-1} (\sbm x)}
=
\int {d [ T(\bm y) ]}
\frac{q(\bm y )}{
|\det \nabla_{\sbm y} T(\bm y)|
}
=
\int d\bm y \ q(\bm y)
=1.
\end{align}

For more detail, see the chapter 11 in \cite{ villani2008optimal} for example.

\subsection{Proof of Theorem 2}\label{ProofOfThm}
It is sufficient to show
\begin{align}
V_D(G, D)
\leq
\mathbb{E}_{\sbm x \sim p} \Big[ D(\bm x) \Big]
-
\mathbb{E}_{\sbm y \sim p_G} \Big[ D(\bm y) \Big]
\end{align}
because the $\tilde{D}(\bm x)$ is defined by $\tilde{D}(\bm x) = D(\bm x)/K$.
Below, we show this inequality in each case.

\paragraph{Logistic}
Because of the monotonicity of $\log$, 
\begin{align}
-\log (1 + e^{a} )
\leq
-\log e^a
\end{align}
is satisfied for arbitrary $a\in \mathbb{R}$.
So the objective $V_D$ defined by logistic loss enjoys
\begin{align}
V_D(G, D)
&=
-
\mathbb{E}_{\sbm x \sim p} \Big[
\log \Big(
1 + e^{- D(\sbm x)}
\Big)
\Big]
-
\mathbb{E}_{\sbm y \sim p_G} \Big[
\log \Big(
1 + e^{+D(\sbm y)}
\Big)
\Big]
\notag \\
&\leq 
-
\mathbb{E}_{\sbm x \sim p} \Big[
\log \Big(
e^{- D(\sbm x)}
\Big)
\Big]
-
\mathbb{E}_{\sbm y \sim p_G} \Big[
\log \Big(
e^{+D(\sbm y)}
\Big)
\Big]
\notag \\
&= 
\mathbb{E}_{\sbm x \sim p} \Big[
D(\bm x)
\Big]
-
\mathbb{E}_{\sbm y \sim p_G} \Big[
D(\bm y)
\Big].
\end{align}
\paragraph{Hinge}
On the hinge loss, we use the inequality
\begin{align}
\min(0, u) \leq u
\end{align}
as follows.
\begin{align}
V_D(G, D)
&=
\mathbb{E}_{\sbm x \sim p} \Big[
\min \Big(
0,
-1 + D(\bm x)
\Big)
\Big]
+
\mathbb{E}_{\sbm y \sim p_G} \Big[
\min \Big(
0,
-1 - D(\bm y)
\Big)
\Big]
\notag \\
&\leq 
\mathbb{E}_{\sbm x \sim p} \Big[
-1 + D(\bm x)
\Big]
+
\mathbb{E}_{\sbm y \sim p_G} \Big[
-1 - D(\bm y)
\Big]
\notag \\
&\leq 
\mathbb{E}_{\sbm x \sim p} \Big[
D(\bm x)
\Big]
-
\mathbb{E}_{\sbm y \sim p_G} \Big[
D(\bm y)
\Big].
\end{align}
\paragraph{Gradient penalty}
The objective function for discriminator in WGAN-GP is
\begin{align}
V_D(G, D)
=
\mathbb{E}_{\sbm x \sim p} \Big[
D(\bm x)
\Big]
-
\mathbb{E}_{\sbm y \sim p_G} \Big[
D(\bm y)
\Big]
-
\text{penalty},
\end{align}
and the penalty term is defined by
\begin{align}
\text{penalty}
=
\lambda \mathbb{E}_{\sbm x}
\Big[
\Big|\Big|
\nabla_{\sbm x} D(\bm x) - 1
\Big|\Big|^2
\Big]
\geq 0
\end{align}
for a certain positive value $\lambda$ which immediately gives the inequality.

\newpage
\section{Details on experiments}
\subsection{2d experiment}
\paragraph{Training of GAN}
We use same artificial data used in \cite{petzka2017regularization}.
25 gaussians data is generated as follows.
First, we generate 100,000 samples from $\mathcal{N}(\bm 0, $(1e-2)$\cdot \bm I_{2\times2})$.
After that, we divide samples to 25 classes of 4,000 sub-samples and rearrange their center to $\{ -4, -2, 0, +2, +4\} \times \{ -4, -2, 0, +2, +4\} \subset \mathbb{R}^2$.
To make the data variance 1, we divide all sample coordinates by 2.828.
Swissroll data is generated by scikit-learn with 100,000 samples with noise=0.25.
The swissroll data coordinates are also divided by 7.5.

%
We only use WGAN-GP in this experiment.
The number of update for $D$ is 100 if number of iteration is less than 25 and 10 otherwise per one update for $G$.
We apply Adam with $(\alpha, \beta_1, \beta_2) = ($1e-4$, 0.5, 0.9)$ to both of $G$ and $D$.
Under these setup, we train WGAN-GP 20k times with batchsize 256.
We summarize the structure of our models in Table \ref{tab:2dGAN}.
\gantable{0.6}{
$\bm z \sim U([-1, 1]^2)$ \rule[-5pt]{0pt}{15pt}  \\ \hline
dense $\to$ 256 lReLU \\ \hline
dense $\to$ 256 lReLU \\ \hline
dense $\to$ 256 lReLU \\ \hline
dense $\to$ 2 
}{0.}{0.1}{
2d vector $\bm x \in \mathbb{R}^2$ \\ \hline
dense $\to$ 512 lReLU \\ \hline
dense $\to$ 512 lReLU \\ \hline
dense $\to$ 512 lReLU \\ \hline
dense $\to$ 1
}{
GAN architecture in 2d experiment.
}{2dGAN}
\paragraph{DOT}
First, we calculate the $K_\text{eff}$.
We draw 100 pairs of independent samples $(\bm x, \bm y)$ from $p_G$ for calculating their gradient by $l_2$-norm, and take the maximum gradient as $K_\text{eff}$.
In the experiment, we run 10 independent trials and take mean value.
Actual values of $K_\text{eff}$ are 1.68 for 25 gaussians and 1.34 for swissroll in the experiment.

To apply the target space DOT shown in Algorithm 1, we use Adam optimizer with $(\alpha, \beta_1, \beta_2) = (0.01, 0, 0.9)$ for searching $\bm x$ in both of DOT and Naive transports.
We run the gradient descent 100 times, and calculate the Earth-Mover's distance (EMD) between randomly chosen 1,000 training samples and 1,000 generated samples by each method.
We repeat this procedure 100 times, and get the mean value and std of the EMD.
\paragraph{EMD}
Earth Mover's distance (EMD) can be regarded as a discrete version of the Wasserstein distance.
Suppose $\{ x_i\}_{i= 1,2, \dots, N}$ and $\{ y_i\}_{i= 1,2, \dots, N}$ are samples on $X$.
EMD is defined by
\begin{align}
\text{EMD}(\{x_i\},  \{y_j\})
=
\min_{\pi \in \Pi(\{x_i\}, \{ y_j \})} \sum_{i, j=1}^N \pi_{ij} d(x_i, y_j),
\end{align}
where $\pi$ is constraint on
\begin{align}
\pi_{ij} \in \{ 0, \frac{1}{N}\}, \quad
\sum_{i=1}^N \pi_{ij}= \sum_{j=1}^N \pi_{ij} = \frac{1}{N}.
\end{align}
If we regard samples as discrete approximation of the distribution, EMD measures how two distributions are separated.
So if $x_i$ and $y_j$ are sampled from same distribution, the value is expected to be close to zero.
In our paper, we use python library \cite{flamary2017pot} to calculate it. It used $d(x,y) = ||x - y||_2^2$ by default.

\subsection{Experiments on CIFAR-10 and STL-10}
\paragraph{Training of GAN}
We use conventional CIFAR-10 dataset.
On STL-10, we downsize it to $48 \times 48$ instead of using the original size $96\times96$.
Each pixel is normalized so that it takes value in $[-1, 1]$.

On WGAN and SNGAN, we apply 5 updates for $D$ per 1 update of $G$ and use Adam on $G$ and $D$ with same hyperparameter: $(\alpha, \beta_1, \beta_2) = (0.0002, 0.0, 0.9)$.
We use the gradient penalty on WGAN with $\lambda = 10.0$.
On SAGAN, we apply ``two timescale update rule" \cite{heusel2017gans}, i.e. 1 update for $D$ per 1 update of $G$, and Adam with 
$(\alpha^G, \beta_1^G, \beta_2^G)=(0.0001, 0.0, 0.9)$ and 
$(\alpha^D, \beta_1^D, \beta_2^D)=(0.0004, 0.0, 0.9)$.
Under this setting, we update each GAN 150k times with batchsize 64, except for ResNet SAGAN on STL-10 which is trained by 240k times in the same setup.

We use conventional DCGAN with or without self-attention (SA) layer and normalized layers by spectral normalization (SN) and ResNet including SA layer.
We use usual DCGAN architecture on WGAN and SNGAN used in \cite{miyato2018spectral}.
On SAGAN, we insert a self-attention layer on the layer with 128 channels because it enjoys the best performance within our trials.
We show our DCGAN model in Table \ref{tab:DCGAN} and ResNet in Table \ref{tab:ResNet}.

\gantable{0.4}{
$\bm z \sim U([-1, 1]^{128})$ \rule[-5pt]{0pt}{15pt}  \\ \hline
(SN)dense $\to$ BN $\to M_g \times M_g \times 512$ \rule[-5pt]{0pt}{15pt} \\ \hline 
$4 \times 4$, str=$2$, pad=$1$, (SN)deconv. BN $256$ ReLU \rule[-5pt]{0pt}{15pt}  \\ \hline
$4 \times 4$, str=$2$, pad=$1$, (SN)deconv. BN $128$ ReLU \rule[-5pt]{0pt}{15pt}  \\ \hline
(128 SA with 16 channels) \rule[-5pt]{0pt}{15pt}  \\ \hline
$4 \times 4$, str=$2$, pad=$1$, (SN)deconv. BN $64$ ReLU \rule[-5pt]{0pt}{15pt}  \\ \hline
$3 \times 3$, str=$1$, pad=$1$, (SN)deconv. $3$ Tanh \rule[-5pt]{0pt}{15pt} 
}{0.12}{0.4}{
%
RGB image $\bm x \in [-1, 1]^{M \times M \times 3}$ \rule[-5pt]{0pt}{15pt}  \\ \hline
$3\times3$, str=$1$, pad=$1$, (SN)conv 64 lReLU \rule[-3pt]{0pt}{13pt}  \\
$4\times4$, str=$2$, pad=$1$, (SN)conv 128 lReLU \rule[-5pt]{0pt}{15pt}  \\ \hline
(128 SA with 16 channels) \rule[-5pt]{0pt}{15pt}  \\ \hline
$3\times3$, str=$1$, pad=$1$, (SN)conv 128 lReLU \rule[-3pt]{0pt}{13pt}  \\
$4\times4$, str=$2$, pad=$1$, (SN)conv 256 lReLU \rule[-5pt]{0pt}{15pt}  \\ \hline
$3\times3$, str=$1$, pad=$1$, (SN)conv 256 lReLU \rule[-3pt]{0pt}{13pt}  \\
$4\times4$, str=$2$, pad=$1$, (SN)conv 512 lReLU \rule[-5pt]{0pt}{15pt}  \\ \hline
$3\times3$, str=$1$, pad=$1$, (SN)conv 512 lReLU \rule[-5pt]{0pt}{15pt}  \\ \hline
(SN)dense $\to$ 1
}{
DCGAN model. 
In WGAN, we use bare dense, deconv, conv without self-attentions in both generator and discriminator.
In SNGAN, we use SNdense and SNconv in discriminator but bare dense and deconv in generator without self-attentions. 
In SAGAN, all SN and SA are turned on.
We use the SA layer defined in Figure \ref{tab:SA}.
We use $(M_g, M)=(4, 32)$ in CIFAR-10, $(M_g, M)=(6, 48)$ in STL-10.
}{DCGAN}

\begin{figure}[h]
 \centering
 \hspace{0pt}
\xymatrix@C=0pt@R=10pt{
\text{(width=$w$, hight=$h$, channels=$c$)}
\ar[ddd]
\\
\ar@{->}[r] 
\ar@/^40pt/[rr] 
& 
 \fbox{
\parbox{90pt}{
\centering
``key'' \\
$1 \times 1$, 1, 0 SNconv $s$
\\
$\to $ ($w*h, s$)
}
}
 \ar@{->}[rd] &
\fbox{
\parbox{90pt}{
\centering
``query'' \\
$1 \times 1$, 1, 0 SNconv $s$
\\
$\to $ ($w*h, s$), transpose $\to (s, w*h)$
}
}
\ar@{->}[d]
 &
\\
\ar@{->}[r] & 
\fbox{
\parbox{90pt}{
\centering
``value'' \\
$1 \times 1$, 1, 0 SNconv $c$
\\
$\to $ ($w*h, c$)
}
}
\ar@{->}[d]
&
\fbox{
\parbox{90pt}{
\centering
matmul, \\
softmax along \\
``key'' vectors, \\
transpose,
\\
$\to $ ($w*h, w*h$)
}
}
\ar@{->}[ld]
\\
\oplus
 \ar@{->}[d] & 
 \fbox{
\parbox{90pt}{
\centering
matmul
$\to$ ($w, h, c$) 
}
}
\ar@{->}[l]^{\times \beta}
&&
\\
&
&
}
\caption{$c$ self-attention with $s$ channels.}
\label{tab:SA}
\end{figure}
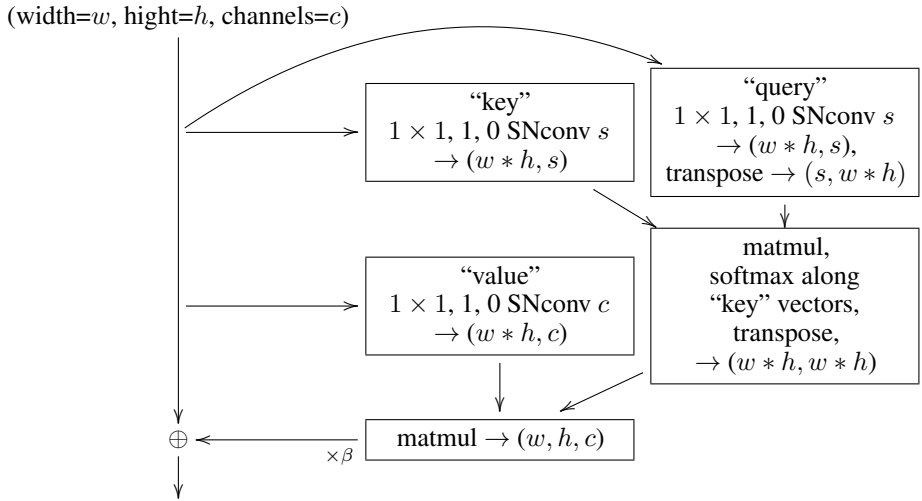

\gantable{0.4}{
$\bm z \sim N(0, \bm I_{128 n\times 128 n})$ \rule[-5pt]{0pt}{15pt}  \\ \hline
%
%
SNdense $\to$ ReLU $M_g \times M_g \times 128n$\rule[-5pt]{0pt}{15pt}  \\ \hline
SNResBlock up 128n\rule[-5pt]{0pt}{15pt}  \\ \hline
SNResBlock up 128n\rule[-5pt]{0pt}{15pt}  \\ \hline
SNResBlock up 128n\rule[-5pt]{0pt}{15pt}  \\ \hline
128n SA with16n channels\rule[-5pt]{0pt}{15pt}  \\ \hline
ReLU, BN, $3 \times 3$ (SN)conv, 3 Tanh \rule[-5pt]{0pt}{15pt}  
}{0.12}{0.4}{
%
RGB image $\bm x \in [-1, 1]^{M\times M\times 3}$ \rule[-5pt]{0pt}{15pt}  \\ \hline
SNResBlock down1 128 \rule[-5pt]{0pt}{15pt}  \\ \hline
SNResBlock down2 128 \rule[-5pt]{0pt}{15pt}  \\ \hline
SNResBlock down3 128 \rule[-5pt]{0pt}{15pt}  \\ \hline
SNResBlock down3 128 \rule[-5pt]{0pt}{15pt}  \\ \hline
128 SA with 16 channels \\ \hline
ReLU, SNdense $\to 1$
}{
ResNet model. In this paper, we concentrate on ResNet with spectral normalization and self attention trained by CIFAR-10 (n, $M_g, M$)=($2, 4, 32$), SLT-10(n, $M_g, M$)=($1, 6, 48$).
See Figure \ref{tab:??} and Figure \ref{tab:!!} for definitions of ResBlocks.
}{ResNet}

\begin{figure}[th]
 \begin{minipage}{0.4 \hsize}
 \small
 \centering
\xymatrix@C=0pt@R=10pt{
\ar[dd]
\\
\ar@{->}[r] & \fbox{\text{BN, ReLU} }\ar@{->}[d]
\\
\fbox{\text{$2\times2, 2, 0$ unpooling}}  \ar@{->}[d] & 
\fbox{\text{$2\times2, 2, 0$ unpooling}}  \ar@{->}[d] 
\\
\fbox{
\parbox{83pt}{
\centering
$3 \times 3$, 1, 1 (SN)conv, \\ ReLU}
}
 \ar@{->}[d] & 
\fbox{\text{BN, ReLU}} \ar@{->}[d] 
\\
\oplus \ar[d]&
\fbox{
\parbox{83pt}{
\centering
$3 \times 3$, 1, 1 (SN)conv, \\ ReLU}
}
\ar@{->}[l] 
\\
 &
}
\hspace{70pt}
(SN)ResBlock up
\end{minipage}
\begin{minipage}{0.\hsize}
\
\end{minipage}
 \begin{minipage}{0.5\hsize}
 \small
 \centering
\xymatrix@C=0pt@R=10pt{
\ar[dd]
\\
\ar@{->}[r] & 
\fbox{\text{$3 \times 3$, 1, 1 (SN)conv}} \ar@{->}[d]
\\
 \fbox{\text{$4\times4, 2, 1$(SN)conv}}   \ar@{->}[d] & 
 \fbox{\text{ReLU}} \ar[d]
 \\
 \oplus \ar[d]
 &
 \fbox{\text{$4\times4, 2, 1$(SN)conv} } \ar[l]
\\
 &
}
(SN)ResBlock down1
\end{minipage}
 \caption{ResBlocks}
 \label{tab:??}

\vspace{20pt}
 \begin{minipage}{0.4 \hsize}
 \small
 \centering
\xymatrix@C=0pt@R=10pt{
\ar[dd]
\\
\ar@{->}[r] & 
\fbox{
\parbox{90pt}{
\centering
ReLU, \\
$3 \times 3$, 1, 1 (SN)conv}
}
 \ar@{->}[dd]
\\
\fbox{\text{$4\times4, 2, 1$ (SN)conv}}  \ar@{->}[d] & 
\\
\oplus
 \ar@{->}[d] & 
 \fbox{
 \parbox{90pt}{
\centering
ReLU, \\
$4\times4, 2, 1$ (SN)conv} 
}
\ar@{->}[l] 
\\
&
}
\hspace{70pt}
(SN)ResBlock down2
\end{minipage}
\begin{minipage}{0.1\hsize}
\
\end{minipage}
 \begin{minipage}{0.35\hsize}
 \small
 \centering
\xymatrix@C=30pt@R=10pt{
\ar[ddd]
\\
\ar@{->}[r] & 
\fbox{
\parbox{90pt}{
\centering
ReLU, \\
$3 \times 3$, 1, 1 (SN)conv}
}
 \ar@{->}[dd]
\\
& 
\\
\oplus
 \ar@{->}[d] & 
 \fbox{
 \parbox{90pt}{
\centering
ReLU, \\
$3\times3, 1, 1$ (SN)conv} 
}
\ar@{->}[l] 
\\
&
}
(SN)ResBlock down3
\end{minipage}
\caption{ResBlocks}
\label{tab:!!}
\end{figure}

\newpage
\paragraph{DOT}
First of all, let us pay attention to the implementation of SN proposed in \cite{miyato2018spectral}.
The algorithm gradually approximate SN by Monte Carlo sampling based on forward propagations, and does not give well normalized weights in the beginning, so we should be careful to apply DOT on such network.
One easy way is just running forward propagations a few times.
Before each DOT, we run forward propagation on $G$ and $D$ to thermalize the SN layers.
We apply SGD update with\footnote{
lr corresponds to $\epsilon$ in the main paper.
} lr=0.01.
In Table 1 of the main paper, we update each generated samples with 20 times for DCGAN, 10 times for ResNet.

To get $k_\text{eff}$, we draw 100 pairs of samples $(\bm x, \bm y)_i$, calculate maximum gradient, and define it as $k_\text{eff}$.
However, there seems no big difference to use $k_\text{eff}$ in high precision or not.
To compare them, we executed $k_\text{eff}=1$ DOT and summarize scores (IS, FID) on 0, 10 and 20 updates.
\begin{align}
&
\left. \begin{array}{l|l|l|l|l}
& 
\text{\# updates}=0 & 
\text{\# updates}=10 & 
\text{\# updates}=20
\\ \hline
\text{trial1}(k_\text{eff}=1.00)
& 6.47(05), 27.83
& 7.17(07), 24.31
& 7.35(01), 24.06
\\
\text{trial2}(k_\text{eff}=0.86)
& 6.53(08), 27.84
& 7.21(01), 24.06
& 7.45(05), 24.14
\end{array} \right.
\notag
\\
&
\hspace{30mm}
\text{WGAN-GP(CIFAR-10, lr=0.01)}
\notag
\\ 
&
\left. \begin{array}{l|l|l|l|l}
\text{Number of updates} & 
\text{\# updates}=0 & \text{\# updates}=10 & \text{\# updates}=20
\\ \hline
\text{trial1}(k_\text{eff}=1.00)
& 7.44(01), 20.71
& 7.63(05), 18.38
& 7.69(09), 17.74
\\
\text{trial2}(k_\text{eff}=0.39)
& 7.45(09), 20.74
& 7.85(08), 16.57
& 7.97(14), 15.78
\end{array} \right.
\notag
\\
&
\hspace{30mm}
\text{SNGAN(ns)(CIFAR-10, lr=0.01)}
\notag
\\ 
&
\left. \begin{array}{l|l|l|l|l}
\text{Number of updates} & 
\text{\# updates}=0 & \text{\# updates}=10 & \text{\# updates}=20
\\ \hline
\text{trial1}(k_\text{eff}=1.00)
& 7.4(01), 20.32
& 7.6(07), 19.3
& 7.61(08), 19.01
\\
\text{trial2}(k_\text{eff}=0.34)
& 7.45(08), 20.47
& 7.81(08), 17.72
& 8.02(16), 17.12
\end{array} \right.
\notag
\\
&
\hspace{30mm}
\text{SNGAN(hi)(CIFAR-10, lr=0.01)}
\notag
\\ 
&
\left. \begin{array}{l|l|l|l|l}
\text{Number of updates} & 
\text{\# updates}=0 & \text{\# updates}=10 & \text{\# updates}=20
\\ \hline
\text{trial1}(k_\text{eff}=1.00)
& 7.66(07), 25.09
& 7.92(14), 23.1
& 8.02(12), 22.48
\\
\text{trial2}(k_\text{eff}=0.28)
& 7.75(07), 25.37
& 8.35(11), 21.27
& 8.5(01), 20.57
\end{array} \right.
\notag
\\
&
\hspace{30mm}
\text{SAGAN(ns)(CIFAR-10, lr=0.01)}
\notag
\\ 
&
\left. \begin{array}{l|l|l|l|l}
\text{Number of updates} & 
\text{\# updates}=0 & \text{\# updates}=10 & \text{\# updates}=20
\\ \hline
\text{trial1}(k_\text{eff}=1.00)
& 7.46(01), 26.08
& 7.75(11), 24.12
& 7.87(09), 23.33
\\
\text{trial2}(k_\text{eff}=0.21)
& 7.52(06), 25.78
& 8.2(08), 21.45
& 8.38(05), 21.21
\end{array} \right.
\notag
\\
&
\hspace{30mm}
\text{SAGAN(hi)(CIFAR-10, lr=0.01)}
\notag
\end{align}
As one can see, lower $k_\text{eff}$ makes improvement faster.
But please note that if it is too small, the DOT may be equivalent just decreasing $-D(\bm x)$, and easily increase FID.


%
On the lr of gradient decent, it is better to take small value as possible.
For example, the history of DOT for ResNet on STL-10 is as follows.
\begin{align}
\left. \begin{array}{l|c|c|c|}
& 
\text{\# updates}=0 & \text{\# updates}=10 & \text{\# updates}=20
\\ \hline
\text{Inception score} & 
9.33(08) & 10.03(14) & 10.00(12)
\\ \hline
\text{FID} & 
41.91 & 39.48 & 40.53
\\ \hline
\end{array} \right.
\notag
\end{align}
In this model we use $\mathcal{N}(0, \bm I_{128\times 128})$ as the prior $p_Z$ and apply the projection of the gradient to conduct DOT updates.
But our projection update is an approximation, and the slightly bad scores on 20 updates may be caused by $\bm z$ getting out of the support because of too large lr.
On the other hand, our DCGAN model has $U[-1,1]^{128}$ as the prior, and there is no need of the projection.
In this case, for example, SAGAN(hi)'s history is
\begin{align}
\left. \begin{array}{l|c|c|c|}
& 
\text{\# updates}=0 & \text{\# updates}=10 & \text{\# updates}=20
\\ \hline
\text{Inception score} & 
9.29(13) & 10.11(14) & 10.29(21)
\\ \hline
\text{FID} & 
45.78 & 41.09 & 40.51
\\ \hline
\end{array} \right.
\notag
\end{align}
and each score improved even after 10 updates.
We show some results on DOT histories with different lr in Figure \ref{fig:kS}, Figure \ref{fig:kL}, Figure \ref{fig:kL2} also. 
As we can see from Figure \ref{fig:kS}, larger lr makes improvement faster, e.g. IS 7.40 reaches 8.88 and FID 22.37 reaches 17.61 at 10 update point, but it easily makes bad scores when we increase the number of updates, e.g. IS 8.88 at 10 reaches 7.79 and FID 17.61 at 10 reaches 58.64 at 90 update point.
This is resolved by taking lower lr, but too low lr makes improvement too slow as we can see in Figure \ref{fig:kL2}.

\paragraph{Inception score and FID}
Inception score is defined by
\begin{align}
IS(\{\bm x_i \}_{i})
=
\exp{
\sum_{i=1}^N \frac{1}{N}
\Big(
\hat{D} ( p(\bm d|\bm x_i) || p(\bm d) )
\Big)
},
\end{align}
where $p(\bm d|\bm x)$ is the output values of the inception model, and $p(\bm d)$ is marginal distribution of $p(\bm d|\bm x) p_G(\bm x)$.
This is one of well know metrics on GAN, and measures how the images $\{ \bm x_i \}_i$ look realistic and how the images have variety.
Usually, higher value is better.

The second well know metric is the Fr\'echet inception distance (FID).
This value is the Wasserstein-2 distance between dataset and $\{ \bm x_i \}_{i=1,2, \dots, N}$ in the 2,048 dimensional feature space of the inception model by assuming the distribution is gaussian.
To compute it, we prepare the 2,048 dimensional mean vector $\bm m_w$ and covariant matrix $\bm C_w$ of the corresponding dataset, and calculate $\bm m$ and $\bm C$ by feeding $\{ \bm x_i\}_{i=1,2, \dots, N}$ to the inception model.
Then, the FID is calculated by
\begin{align}
\text{FID}(\{ \bm x_i \}_{i})
=
|| \bm m - \bm m_w ||_2^2
+
\text{Tr}(\bm C + \bm C_w - 2(\bm C \bm C_w)^{1/2} ).
\end{align}
Note that the square root of the matrix is taken under matrix product, not component-wise root as usually taken in numpy.
Lower FID is better.

\paragraph{DOT vs Naive}
Here, we compare the latent space DOT and the latent space Naive improvement:
\begin{align}
T^\text{naive}_{D \circ G}(\bm z_{\sbm y} ) =
\text{argmin}_{\sbm z}
\Big\{
- \frac{1}{k_\text{eff}} D \circ G(\bm z)
\Big\} 
.
\label{naiveL}
\end{align}
As one can see from Figure \ref{fig:kL} and Figure \ref{fig:kL2}, both the DOT and the naive transport \eqref{naiveL} improve scores.
In Figure \ref{fig:kL},DOT and Naive keep improving the inception score, on the other hand, the FID seems saturated around 40$\sim$50 updates.
After that, one can see both of transports do not improve FID.
Even worse, FID starts to increase both cases at some point of updates.
Compared to the naive update, however, DOT can suppress it, but increasing FID at some update point seems inevitable. So, keeping lr low value as possible seems important as we have already noted.

\begin{figure}[h]
\centering
\includegraphics[width=400pt]{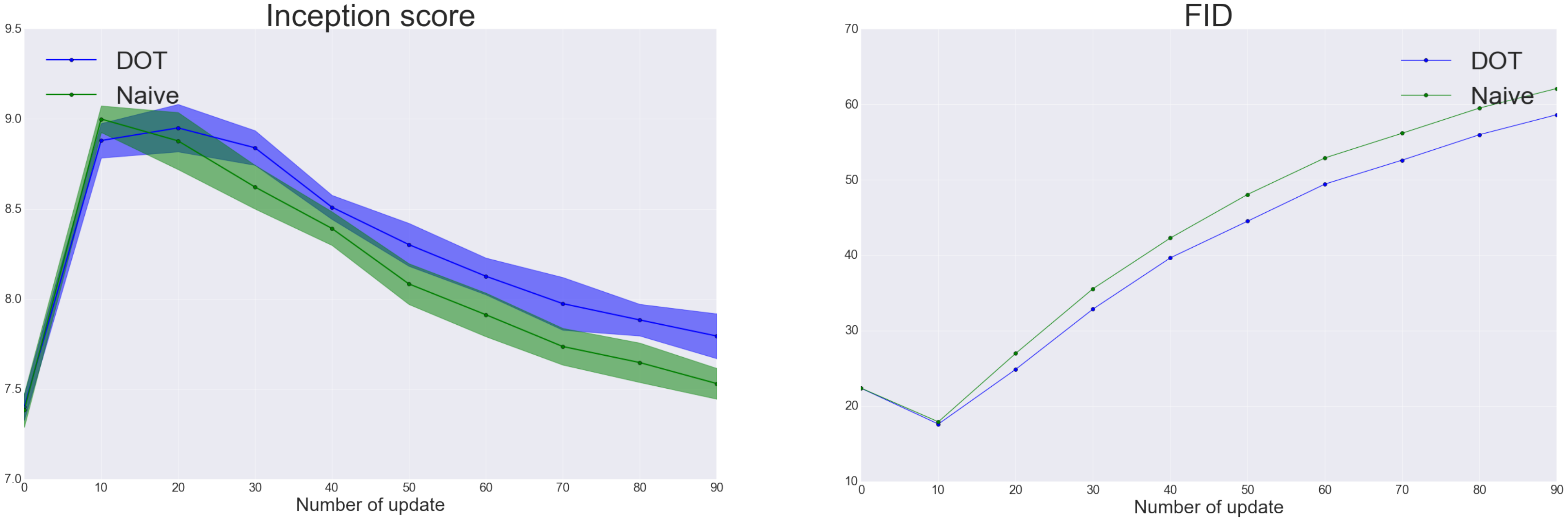}
\vspace*{-2mm}
\caption{
History of inception scores and FID during each transport with $lr=0.05, k\approx0.29$ with SNGAN model trained by CIFAR-10.
Too large lr causes bad behavior.
}
\label{fig:kS}

\vspace{10mm}
\includegraphics[width=400pt]{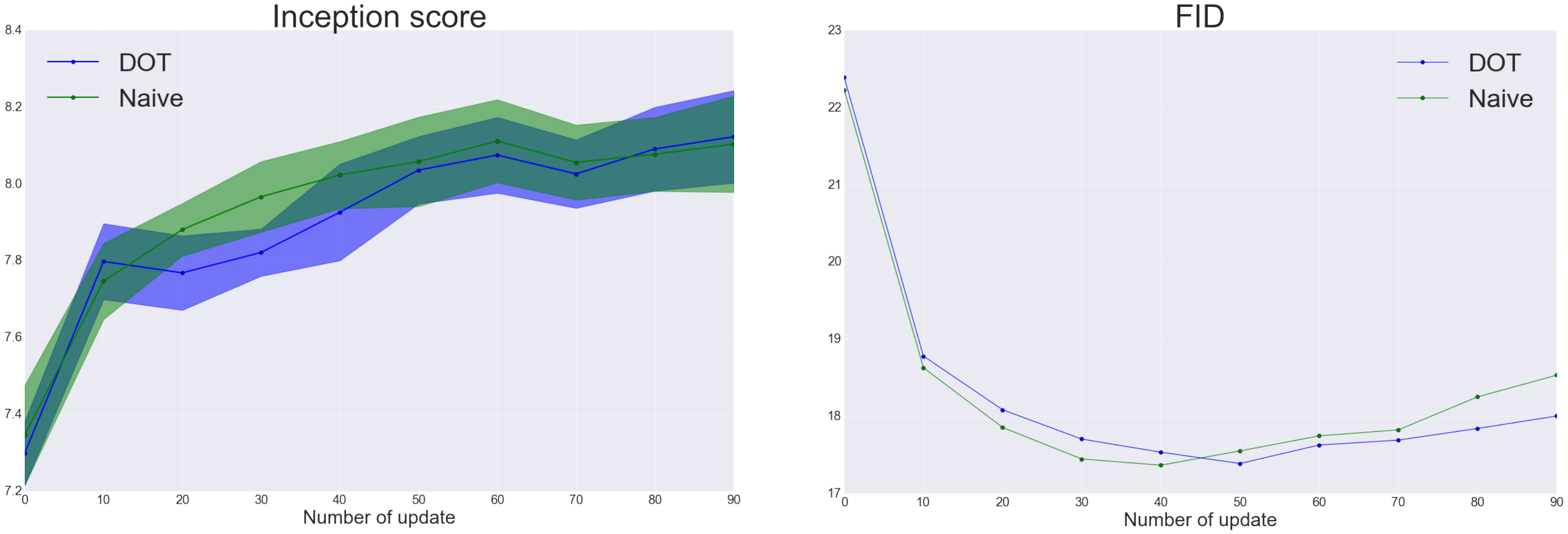}
\vspace*{-2mm}
\caption{
History of inception scores and FID during each transport with $lr=0.005, k\approx0.31$ with SNGAN model trained by CIFAR-10.
}
\label{fig:kL}

\vspace{10mm}
\includegraphics[width=400pt]{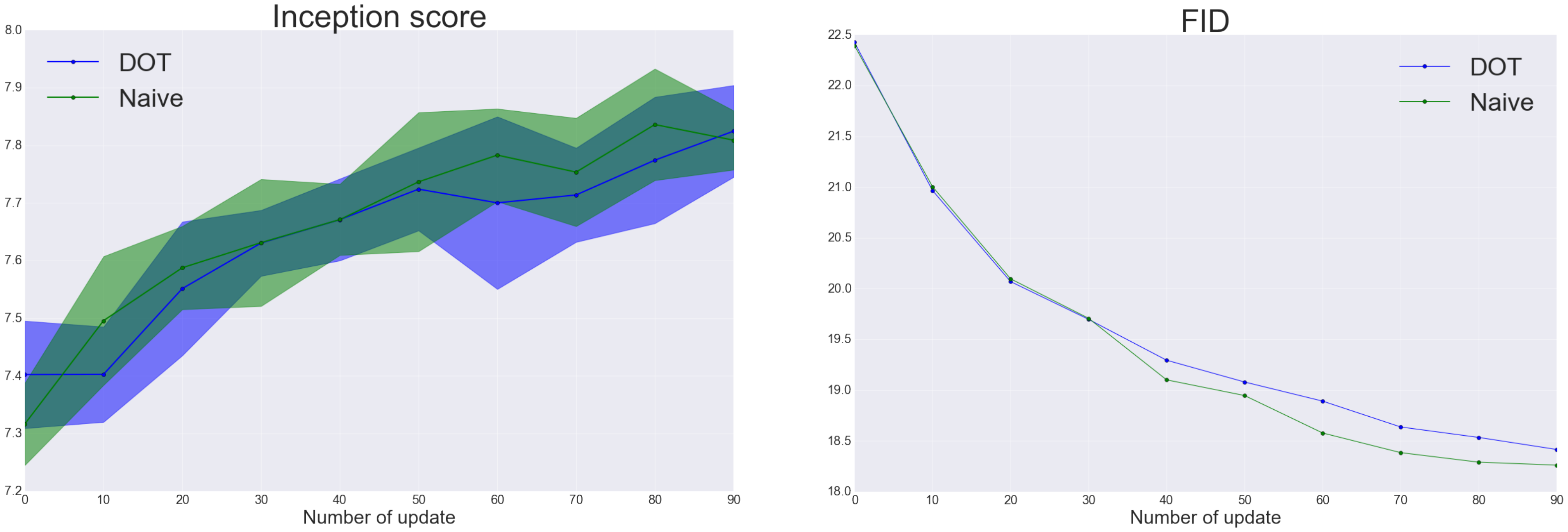}
\vspace*{-2mm}
\caption{
History of inception scores and FID during each transport with $lr=0.001, k\approx0.28$ with SNGAN model trained by CIFAR-10.
Too low lr causes slowing down the speed of improvement.
}
\label{fig:kL2}
\end{figure}
\paragraph{DOT vs MH-GAN}
There are some methods of post-processing using trained models of GAN \cite{azadi2018discriminator, turner2019metropolis}.
In this section, we focus on the Metropolis-Hastings GAN (MH-GAN) \cite{turner2019metropolis} which is relatively easy to implement.
In MH-GAN, we first calibrate the trained discriminator by logistic regression, and use it as approximator of the accept/reject probability in the context of the Markov-Chain Monte-Carlo method for sampling.
We calibrate $D$ by using $1,000$ training data and $1,000$ generated data, and run MC update 500 times.
 \begin{table}[h]
\centering
  \begin{tabular}{l|cc|cc|}
 & \multicolumn{2}{|c|}{CIFAR-10}
 & \multicolumn{2}{|c|}{STL-10}
 \\ \hline
   & bare & MH-GAN 
      & bare & MH-GAN \\
 \hline
WGAN-GP
& 6.5(08), 27.93
& 7.23(11), 36.14
& 8.71(13), 49.98
& {\bf 8.98(13), 48.03}
\\
SNGAN(ns)
& 7.42(09), 20.73
& 7.16(01), 23.24
& 8.62(15), 41.35
& 8.0(11), 46.27
\\
SNGAN(hi)
& 7.44(08), 20.53
& {\bf 8.23(12), 18.57}
& 8.78(01), 40.11
& {\bf 10.02(08), 36.34}
\\
SAGAN(ns)
& 7.69(08), 24.97
& {\bf 7.87(07), 22.48}
& 8.63(08), 48.33
& {\bf 9.79(12), 44.44}
\\
SAGAN(hi)
& 7.52(06), 25.77
& {\bf 7.92(09), 23.75}
& 9.32(11), 45.66
& 9.73(19), 49.1
\\ \hline
  \end{tabular}
 \caption{(Inception score, FID) by usual sampling (bare) and MH-GAN in within our DCGAN models. The bold letter scores correspond increasing inception score and decreasing FID.
 }
 \label{tab:scoresmh}
\end{table}

We succeed in improving almost all inception scores except for SNGAN(ns) cases.
On FID, however, MH-GAN sometimes downgrade it (taking higher value compared to its original value).
By comparing Table \ref{tab:scoresmh} and Table 1 in the main body of this paper, DOT looks better in all cases, but we do not insist our method outperform MH-GAN here because our DOT method needs tuning parameters $\epsilon, k_\text{eff}$ besides tuning the number of update.

\subsection{On runtimes}
We just used gradient of $G$ and $D$, so it scales same as the backprop. 
For reference, we put down real runtimes (seconds/30updates) here by Tesla P100:
\begin{align}
\left. \begin{array}{|l|l|l|l|l|}
\hline
 \text{Swissroll} &
 \text{CIFAR-10 (SN-DCGAN)} &
  \text{STL-10 (SN-DCGAN)} &
   \text{ImageNet} 
 \\
 \hline
0.310(02) 
 &
1.04(01) 
 &
1.05(01) 
 &
2.52(01) 
 \\
 \hline
\end{array} \right.
\notag
\end{align}
The error is estimated by 1std on 10 independent runs.

\input{endbib}
\bibliographystyle{plain}

\end{document}